# Unified vector space mapping for knowledge representation systems

Dmytro Filatov,
Taras Filatov  PhD.

**Abstract**

One of the most significant problems which inhibits further developments in the areas of Knowledge Representation and Artificial Intelligence is a problem of semantic alignment or knowledge mapping. The progress in its solution will be greatly beneficial for further advances of information retrieval, ontology alignment, relevance calculation, text mining, natural language processing etc. In the paper the concept of multidimensional global knowledge map, elaborated through unsupervised extraction of dependencies from large documents corpus, is proposed. In addition, the problem of direct Human – Knowledge Representation System interface is addressed and a concept of adaptive decoder proposed for the purpose of interaction with previously described unified mapping model. In combination these two approaches are suggested as basis for a development of a new generation of knowledge representation systems.

Keywords: *knowledge representation, knowledge mapping, human computer interaction, ontology alignment, upper ontology, relevance calculation, information retrieval, document similarity, search, big data, brain computer interfaces, digital telepathy*

## 1. Introduction

In society, the field of knowledge representation became more significant over last years [25]. People always made attempts to study and classify the knowledge about knowledge. We can find references from as early as Socrates in fifth century B.C. [33] to the flourishing of logic and epistemology [28] in middle ages. Since the problem was considered to be important in past, it is hard to overestimate its meaning during the era of information.

Modern technologies have endowed mankind with excessive floods of data which are difficult to systematise and process. For a person to become a specialist in a certain area it takes years of learning and requires a subsequent informational race to keep up with the latest professional trends.

It is a popular belief among knowledge engineers and data mining specialists that information available in open access is enough to extract truthful facts about virtually any aspect of our life and even predict future. The only problem to resolve is to intelligently process the information from multitude sources. [35].

This factors demand new generation of knowledge representation systems to help mankind systematise, access and use its collective knowledge.

In current paper we propose ideas for development of a new age Knowledge Representation System (KRS).

We believe that recent achievements in certain areas and sciences will soon lead to a tremendous breakthrough in the scope of human knowledge representation and human-computer interaction. This will open new horizons and greatly increase effectiveness of human work in many applications. The only thing that needs to be done is to bring these achievements together.

Ideal KRS should provide a user with a convenient access to all knowledge of mankind. Its elements therefore are: a human who wants to access some piece of knowledge he/she is interested in, a data storage, and the intermediary system to provide interface for a human to access knowledge.
The obstacles which arise here are caused by limitations of human abilities and current technological level.

## 2. Data storage

### 2.1 State of the art

Knowledge representation system requires data storage unless it is able to retrieve necessary documents from external sources in real time. We can outline two winning approaches for an ideal KRS of nowadays: structured manually managed global knowledge storages such as ontologies and systems involving automated indexing and retrieval of most full and accessible raw documents collection (World Wide Web) such as search engines. The problem of the first approach is in its manual nature – any attempts to create and maintain the global human knowledge base will result in compromise between the detail, actuality and usability. Notwithstanding there are multiple successful upper-ontology projects such as Cyc [23], WordNet [12], DnS, SUMO etc and ongoing theoretical discussions as long as research aimed to elaborate a standardised unified global ontology under the names of SUO (Standard Upper Ontology) [27], UFO (Unified Framework Ontology) etc.
The second approach became historically prevalent due to WWW being the largest, comprehensive and up-to-date
corpus of data available for automated processing nowadays. However in contrast to first approach, the problems of automated information retrieval play a significant role here. The problems of text understanding and natural language processing are one of the most challenging in AI and nevertheless still remain without efficient solution. Adjacent are the problems of classification and relevance calculation, the so-called 'web clustering' problem [1]. The second approach (automated indexing) therefore has major lacks in the accuracy of retrieval.

There are ongoing efforts on bringing improvements to the abovementioned approaches in order to overcome these problems. For example, along with unified base ontology projects (SUO, BULO) there are certain attempts to develop ontology alignment and ontology mapping techniques in order to bring existent ontologies together with each other and with other kinds of knowledge bases [5, 17]. It is often proposed to decrease the shortcomings of manual administration in case of ontologies by the means of automated information retrieval (search engine technologies). From the other side, improved hypertext standards are being developed in order to make possible to manually specify information to assist search engines understand the commonsense meaning of the WWW documents and hyperlinks between them [24]. It is necessary to understand that these hybrid solutions carry the shortcomings of corresponding techniques along with advantages.

One shortcoming unites these approaches and makes impossible their integration: there is no standard of mapping and establishing relations between documents and concepts in different systems. The problem would be resolved in case one of the systems with fixed mathematically interpretable hierarchy such as ontologies overcomes the existing approach (WWW). However this seems unlikely due to abovementioned reasons. Independent intermediary standard is a potential solution to this problem. There are multiple initiatives on direction of linking and reciprocal mapping of knowledge bases of different types among which there is an area of ontology mapping. The initiatives have one common shortcoming: no single standard of mapping and establishing relations between documents. Subsequently, none of them is likely to become a widely recognized standard unless a more sustainable solution is developed.

### 2.2 The concept of a Global Knowledge Map (GKM)

We believe it is possible to elaborate a single standard for knowledge mapping by building a logical space with the purpose of projection of real world knowledge concepts. Such model (let us call it

Global Knowledge Map) should reflect level of similarity of documents and concepts mapped onto it. Main purpose of the model is:

- Alignment and cross-mapping of documents and concepts (WWW, ontologies, e-libraries, directories etc)
- Information retrieval through browsing
- Precise automated calculation of commonsense relevance

GKM therefore requires a mathematical/logical model of the knowledge storage with a specific condition: being optimal for the task of knowledge representation i.e. interaction with human. For the fulfillment of this condition the model must reflect in its dimensionality or in its structure the structure of the human knowledge.

The requirements therefore are:

**Dimensionality and mapping**.
The main factor for the dimensionality is the meaning (or a topic).
1.        Each concept of the human knowledge may be mapped onto a point with a specific coordinates in the system space.
2.        Each document or text may be mapped onto a number of points (document is divided into memes - meaningful pieces) or one point.
Relevance calculation.
1.        It is possible mathematically to calculate relevance between two concepts by calculating the distance between their corresponding projection points in the space.
2.        It is therefore possible to calculate 'similarity' between documents and concepts by calculating the distance between their mappings.

**Homogeneity of the space**.
1.        The space is uniform (homogenous) and continuous
2.        Coordinates reflect the meaning and the distances between the points reflect the difference in meaning so that if the point C is situated between A and B then it means concept C is related to both A and B equally.
3.        It is possible to 'browse' the space finding the knowledge sources mapped to the neighboring areas.
Building a mathematical model of such space enables the development of Global Knowledge Map. It is not worthwhile trying to elaborate such a model (GKM) in a manual way due to abovementioned reasons of information growth and continual change in human understanding of the world. We believe it is possible to extract dependencies and rules from available corpuses of texts and use these as processors for our mapping purposes.
        The corner stone of our assumptions is that it is generally possible to map various human knowledge subjects onto single space and the distances within the dimensionality of the latter reflect a level of similarity between given subjects. This assumption is based on Johnson-Lindenstrauss Lemma (2.1)

Given $0 < \varepsilon < 1$, a set $X$ of $m$ points in $\mathbf{R}N$, and a number $n > 8 \ln(m) / \varepsilon^2$, there is a linear map $f : \mathbf{R}N \to \mathbf{R}n$ such that

$$(1 - \varepsilon)\|u - v\|^2 \le \|f(u) - f(v)\|^2 \le (1 + \varepsilon)\|u - v\|^2$$

(2.1)

        for all $u, v \quad X$.

stating that a set of n points in high dimensional Euclidian space can be mapped down to an N dimensional Euclidian space such that the distance between any two points changes by only a factor (1 ) [7]. The Vector Space Model commonly used in Information Retrieval and Text Categorization represents documents as high dimensional vectors [31]. These vectors contain certain level (depending on a metric function chosen) of information which is enough to classify the subject of the original document.

The Tychonoff's theorem [26] states that points, representing the properties of objects of one class, should be situated closer to each other in the property space than to points representing the properties of objects of other classes. In our task this means the original vector space of n texts may be projected onto fixed N-dimensional space and using an appropriate algorithm for data compression / dimensionality reduction due to the theorem of compactness [22] the mapping will be achieved where the distances between points represent the relevance of the corresponding documents.

Factors which affect the precision of the mapping:

- representativeness of the metric function and the size of the feature space
- size, dispersity and quality of the documents corpus

Provided the theory is applicable in current conditions it remains to be found which techniques shall be used to elaborate the mapping. We propose to focus on automatic means due to many complications making manual expert-based mapping inapplicable.

## 2.3 Automatic GKM generation through unsupervised extraction (hypothesis)

In order to be useful GKM should contain mappings of significant number of real word (WWW) documents and in its structure represent the common human understanding of the world. It is not worthwhile therefore to consider any manual ways of creation of GKM and filling it with document mappings. The data mining principles should be used to extract dependencies representing the knowledge from the existent corpus of documents available for computer processing and filter out unnecessary data.

There have been numerous attempts for unsupervised extraction of dependencies in texts however it is still a doubtful question whether any technique is capable to provide a sustainable knowledge extraction through the analysis of large documents collection [10, 11, 21, 30].

Let's divide the factors which generally affect the contents of the documents into three categories:

1.      Authors personal factors (feelings and motives to create the text, physical and moral states during the thinking and writing)

2.      Language (the rules etc of the language used to write the document)

3.      Knowledge (certain parts of human knowledge author transmits directly or implicitly through the document)

Let's presume it is possible to process all available text documents of human authors and extract all the dependency rules. In this case the influence of factor 1 will be minimal. The influence of factor 2 is not of much importance due to the following:

a)   documents in multiple languages might be indexed, therefore reciprocally reducing the influence

b)   the language itself reflects human knowledge [18]; so to a certain extent the factor 2 is a subfactor of 3 and even the extraction of their mixture represents a satisfactory achievement

It is theoretically possible then to extract info mostly corresponding to the human knowledge exposed through the available documents. This info transformed to the mapping space will hypothetically provide us with sustainable GKM.

## 2.4 Implementation (experiment)

In the space built each document should be mapped to single coordinate. The 'browsing' of the space or distances comparison should reveal that situation of documents or their clusters reflect their relevance and that it is possible to assign certain topic names to specific coordinates in the space.

Our experiments of using 2 and 3 dimensional Kohonen SOM with a local collection of documents reveal that the distances between projections of documents are not stable throughout the series of launches. This in our opinion is the evidence of the fact that dimensionality of the map is insufficient which conforms to Johnson-Lindenstrauss Lemma mentioned above.

Unfortunately it is impossible to carry out the experiment with a proper dimensionality. For example, according to Johnson-Lindenstrauss Lemma, to map 20,000 documents allowing 10% error it will require 58 dimensions. This requires calculations which are above modern computers' capacity (if doing real-time calculation).

The important thing to mention here is that, while Lemma gives a maximum dimensionality of the mapping space allowing to fulfill the condition of single projection, it is not necessary the minimal effective value. Lemma gives a value for a set of n points i.e. for the worst case, which is not likely to appear in practice. The methods for dimensionality detection should be used to calculate an effective dimensionality of data and therefore determine the correct number of dimensions for the mapping of particular data set. There are known techniques for this, which come from the background of surface reconstruction. One of the latest is the work by S. Cheng, Y. Wang and Z. Wu [4] where dimension detection method through Principal Component Analysis [8] is presented. Thus, the intrinsic dimensionality of data may be extracted. As this value in practice is lower than maximal dimensionality necessary for the projection of the same number of documents in a worst-case theoretical case, this gives a significant reduce in calculation time. Alternatively for the same purpose we can consider the methods based on iterative evaluations. The purpose of dimensionality reduction in our case is to establish an effective mapping where the commonsense similarity between informational sources is expressed through Euclidian distance between projection points. Taking any dimensionality reduction technique based on random selection from the input data set (such as Principal Component Analysis, Self Organising Maps, Sammon reconstruction or triangulation), in case when output dimensionality is less than appropriate, the projections will be unstable and with each recalculation the mapping will be different and the distances between projections will not preserve. We can also assume that when the effective dimensionality of the input manifold is used for the mapping, the distances will preserve with a certain degree of freedom depending on dimensionality reduction method used. An alternative solution therefore is a random selection dimensionality reduction technique combined with incremental dimensionality parallel evaluation of the mapping. Consider, for example, the technique of growing SOMs described in [19, 32]. Aimed to reduce calculation time, the number of nodes is increased dynamically with new nodes being uniformly distributed among the old ones and their weights vectors being set to the means of the neighbouring original weights vectors. Same technique applied to dimensionality (D) would give an opportunity to evaluate each D step by step. Evaluation criteria would be stabilisation of distances between projections. For this purpose two or more self-organising maps may be run in parallel with pairwise distances between certain projections being compared on each iteration. Under iteration here we understand the stage when the dimensionality of the SOM was increased and the network was consequently stabilized with new parameters.

Summarizing the abovementioned we propose the following model for the experimental evaluation of the better approach for establishing the Global Knowledge Map from the collection of text documents.

- The dataset: vector space model to be used (each document represented as vector with features as dimensions and features' ranks as coordinates in corresponding dimensions).
- Feature selection function: mosteffective to be defined.
- Vector size: to be established empirically.

**Data processing and storage**.

A dimensionality reduction technique should be used for the mapping. There are two possible

approaches:

a)      pre-calculate the intrinsic dimensionality and evaluate different dimensionality reduction methods with a known dimensionality of the mapping;

b)      'incremental dimensionality evaluation approach' with few mappings being run in parallel – only methods with random selection of the input data may be used.

  Inputs: documents' feature vectors. Outputs: GKM coordinates. Evaluation:

  1) Commonsense evaluation of correspondence between initial documents and Euclidian distances of their mapping projections.

  2) Stabilization of these pairwise distances between projections through different launches in case random selection technique is used.

## 3.      Interface

### 3.1      Ideal knowledge representation interface

          The interface part of Knowledge Representation System is important when the ideal system is discussed. Both tasks of receiving requests from users and transmitting knowledge back to them are of equal importance with the tasks of data structuring and storage. In current paper we discuss the ways towards Knowledge Representation Systems of a new generation and therefore the issue of interaction is overviewed in order to establish whether it is possible to provide an idealistic interface by the means of a modern technology.

          The interfaces that are used to support interaction of a human user with modern knowledge representation and information retrieval systems are mainly of 'indexing' type, i.e. users have to know exactly what they are looking for and they also have to specify it linguistically. A common example of such interface is a search engine. As we have mentioned above, search engine and the corpus of WWW documents is the most complete and up-to-date knowledge representation system available nowadays, this being the reason of their popularity. At the same time it is known that the 'indexing' interface is not natural to use for humans but it is the only alternative as 'browsing' approaches are being established very poorly [16]. The reason for that is the problem of knowledge mapping and alignment, which doesn't allow automated classification and representation of documents according to their subjects. However, with the problem of unified knowledge mapping space being resolved, new possibilities appear for the construction of improved, more natural interfaces of 'browsing' type.

### 3.2 High dimensionality and visualisation

          Having mentioned that the resulting global mapping space is likely to be n-dimensional where n is high it is necessary to resolve the problem of visual representation. It is possible for humans to imagine 3D space, therefore, the optimal ways of nD->3D representation are to be evaluated. Dimensionality reduction techniques or multiple representation approach via interface might be used.

          It is important that with the help of the unified knowledge mapping space the error is minimized during the calculation of relevance between documents, and, moreover, the retrieval of relevant documents even from other systems becomes a trivial task. For the end user this means once the system has located the topic he/she is interested in, it will never lead user to irrelevant documents.

### 3.3 Information request chain

          When the tasks of subjects mapping and location, relevance calculation and knowledge space browsing are resolved the most important task which remains to resolve is the problem of an initial request. In state-of-the-art information retrieval
systems the following processes are usually being involved when information is requested:

Human part: 1) Imagination –> 2) formation of linguistic constructions -> 3) manual keyboard input

(voice input) –> <u>Machine part:</u> 4) linguistic decoding –> 5) matching and location –> 6) visualization of matching documents

The chain is long and it is obvious that data loss and corruption is being significant due to double linguistic encoding and decoding. Firstly, user has to formulate the cognitive images into short linguistic sentence. Secondly, system has to decode the sentence in order to understand the subject of user's interest. The described chain of initial request causes dissatisfaction of users of modern search engines due to inability of the system to 'understand' the request resulting in tremendous time loss of the end user. The degree of these retrieval errors and inconveniences caused by interface limitation is significant and will greatly minimize the effect of more precise retrieval and location of documents introduced by the unified mapping.

It is therefore necessary to consider, theoretically at least, the possibility of creation of an ideal short-chain human <–> Knowledge Representation System interaction with units causing data loss being eliminated:

<u>Human part:</u> 1) Imagination –> [ request being passed through direct human brain –GKM coordinates converter ] -> 2) matching and location –> 3) visualization of matching documents

We overview the latest achievements in the area of Brain Computer Interfaces (BCI) for this purpose.

### 3.4 Brain-Computer Interfaces

The research on BCI has been going on for more than 30 years and the area is still very young and develops rapidly. Up to a recent moment most significant advances in the area have been made into artificial limb control i.e. motoring functions of the brain [3] and the interpretation and processing of visual signals. These achievements have been verified during multiple experiments involving animal and human subjects. Researchers report successful integration of mechanical or electronic devices when animals or humans learn to control the device with the help of their brain; others report successful transfer and decoding of visual signals [15, 20].

Brain-computer interfaces studies are closely related to the area of functional neuroimaging, where various technologies have been developed to effectively record the states of person's brain through certain physical characteristics. Most productive from the point of view of BCI is a recent neuroimaging technique called Functional Magnetic Resonance Imaging (fMRI) [6, 13]. This technique allows to record the dynamics of blood flow in different brain areas over time and with a high precision. This consequently allows to establish connections between patterns of activation of various brain areas and certain activities and cognitive processes of the human. It is important that this technique, unlike many alternatives, is non-invasive and doesn't involve injections. It is necessary to note however that the fMRI hardware nowadays is still very expensive and cumbrous.

It is significant that experiments show that brain adapts to new conditions. For example, when motoring impulses were used to control a mechanic manipulator or a computer mouse cursor, brain was able to gradually differentiate and learn to control manipulator separately from artificial limb. Lebedev mentions the effect of 'brain plasticity' which potentially allows to incorporate artificial devices into the body representation. [20]

Recent publications in the field of neuroimaging further still discuss the opportunity of detecting the cognitive states [15]. This stipulates the focusing of our attention on the possibility of application of BCI in human-KRS interaction.

It is known that different cognitive states linked with certain real world objects correspond to certain patterns of brain areas activation. Decoding these patterns allows to understand which superimposed oriented stimuli a person is currently attending (where their attention is directed) or in case with visual objects to identify which class of objects the person is imaging (i.e. faces, buildings, furniture) and even the objects' colour and orientation. [15] These processes are complex and far from being understood at the moment. Further studies shall reveal how low-order and high-order brain signals correlate with certain cognitive functions; how the spatial characteristics of the patterns change over time and under various influences; to which extent it is possible to extrapolate the activation patterns of diverse subjects; etc. It is believed however that precise knowledge of 'computations' performed in human brain is not crucial for the construction of relevant BCIs. [20] Common data mining techniques may be applied to extract useful information from various neuroimaging sensors and establish connections with certain cognitive states.

There are though important issues which can seriously affect the success of appliance of BCI in the area of knowledge representation. Two minor problems are generalization across time and the problem of different instances of the same mental state. It is known that brain areas activation patterns of the same mental states may differ over time. Different instances of the same mental state may give modified images as well, depending on contextual variations and other factors. [15] This requires flexible spatial resampling and classification algorithms to be used as suggested by Haynes and Rees. We believe these problems will be resolved upon development of effective techniques.

More dubious question is the problem of extrapolation to novel cognitive states. Haynes and Rees note that the number of possible perceptual or cognitive states is infinite, whereas the number of training categories is necessarily limited. [15] It is crucial therefore that decoder could be trained to generalize experience obtained from small training set to completely new categories. It would be possible by the means of extrapolation if brain activation patterns are actually
arranged in some systematic parametric space. This remains to be found, however, it is believed it is possible at least for some types of mental content [15]. In case abstract shape space for the classification of neural patterns indeed exists it would allow us to theorize on the possibilities of mapping of human brain cognitive states onto Global Knowledge Map described earlier in this paper. Provided this is achieved, the abovementioned "problem of initial request" will be resolved and "ideal human <–> KRS" chain will be possible to establish.

## 3.5 Learnable Decoder

As now it is known thanks to the latest achievements of brain imaging that it is possible to distinguish the activation of different brain areas when the person is thinking about different subjects we may presume that it is possible to create a learnable decoder to map human initiated cognitive states onto knowledge map of a Knowledge Representation System. Therefore an ideal way of human-computer interaction might be established allowing a tremendous speed and precision of communication with a system. There will be less data loss due to elimination of linguistic stage of interaction. The speed and effectiveness of interaction will increase consequently. These two factors will allow people of various professions to increase the effectiveness of their work significantly. From [6, 15] we know that there are certain regularities of location of brain impulses and the subjects of knowledge, which are common for all humans; we can call these features anthropogenic. However it is known that majority of these links ought to be individualistic. Therefore the decoder must be individually adaptive.

It is also obvious that the efficiency of the decoder will depend on individuals and their training with it and ability to learn. We can presume this from the experiments with artificial interfaces being used to replace lost limbs. Humans and animals were able to concentrate mentally in a special way to move an artificial manipulator and even learn to control the real limb and artificial one separately [20]. Considering the abovementioned we believe that an artificial neural network – based mechanism is the best solution of a decoder problem.
Decoder's learning process

1)   The point with random coordinates in the multidimensional space of GKM is selected.
2)   Multiple documents having their mappings in the neighboring area (Euclidian metric is being used) are selected and displayed to a human operator.
3)   Operator concentrates his/her mind to cognitively attend the given topic and related objects in the memory.
4)   The neuroimaging data is being collected by fMRI hardware over a specific period of time.
5)   The data is processed through a spatial resampling and noise reduction algorithm aimed to extract informative patterns characterizing the current iteration of training.
6)   Prepared data are fetched to the inputs of the neural network. The GKM coordinates of a selected point are fetched to the outputs therefore training the neural network to associate specific brain activation patterns with GKM coordinates.

In such way an individualistic decoder may be trained not only for human <-> KRS interaction but basically human <-> any mechanism interaction. It is known [6, 14] that, there are certain anthropogenic regularities of brain mapping, i.e. in our case it is possible to generalize the linkage of neuroimaging patterns with GKM coordinates over different operators. To make use of it, special 'anthropogenically pre-trained' neural nets may be used. These basic networks are to be prepared through massive collective learning of the same decoder involving a big number of human operators. This will significantly reduce the training time compared to randomly initiated neural network. It might likely occur that it is worthwhile to create different pre-trained decoders for people from different cultural/social/educational clusters. It also remains to be found of how much use the decoder is going to be for immediate use without individual training. It can be presumed that most likely the research and development as well as training of specific decoder models shall be divided into two branches: general/universal that extracts and peruses the patterns typical to all human operators; individual which is focused on extracting and identifying patterns specific to current individual operator, for the benefit of serious improvement in detection and control accuracy. The latter to be aimed at as quick as possible initial training as well as refinement through continuous interaction during normal works process.

## 4. Conclusions

In this paper we have aimed to pursue a target-oriented approach to the problem of research and development of the next generation Knowledge Representation Systems. As a result, innovative concepts have been proposed for both data storage and interface parts of an idealistic KRS.

The concept of the Global Knowledge Map is an idea of multidimensional homogeneous mapping space as an addressing mechanism enabling easy information retrieval and relevance calculation for the information units stored in heterogeneous data warehouses such as WWW, ontologies etc. There have been multiple works on this issue trying to elaborate both visual and semantic mappings of massive documents collections as described in corresponding surveys [2, 9, 29, 34]. However no single concept has found wide application until now. The reasons we believe, along with calculation and implementation difficulties, have roots in the shortcomings of the proposed models. Most mapping models use 2D or 3D space whereas there are theoretical grounds mentioned in this paper which allow us to argue that low dimensional space mapping is not appropriate for real word application. Consequently, here we propose a concept of self-organising multidimensional Global Knowledge Map. The means for automated construction of such unified mapping space are proposed employing the principles of unsupervised extraction and dimensionality reduction techniques. A model for experimental evaluation of described system is proposed.

A possibility of direct human – KRS interface scheme have been concurrently studied. It was revealed that the current stage at which the area of Brain Computer Interfaces potentially allows the construction of such direct chain from the point of view of information request. A concept of learnable decoder applying neuroimaging hardware and neural network based converter is proposed. There are vast possibilities of application of such technology for immediate benefit and further research in the fields of research, communication, knowledge representation and information retrieval (from patent search type tasks to new generation search engines replacing current cumbersome keyword search approach), all sorts of brain controlled computer systems and even digital telepathy as it only remains to pass a unique vector space address (or link) to other person's fMRI transmitter to make the transfer of images and thoughts possible.

The issue of psychological concerns, individual and social impact that might be caused by the technologies proposed was not examined. It is obvious that certain approaches such as brain computer interfaces might, when implemented, violate individual privacy and cause unexpected after-effects. Therefore this is a subject for careful study by researchers in corresponding fields.

There are multiple assumptions and blank spots in the model described. Undoubtedly it must be evaluated through experiments, elaborated and improved with appropriate techniques. This will demand collaborative research and development involving researchers and organisations of various fields. Moreover, there are certain technology barriers to overcome in order to build a described system. Such as: calculation complexity in the case of unsupervised knowledge mapping; a matter of low accessibility and portability of neuroimaging hardware in the case of neuroimaging – global mapping decoder. Nevertheless we believe the ideas presented would be beneficial for researchers working towards elaboration of knowledge representation systems of the next generation.

## References

[1]     J. Allan et al, Challenges in information retrieval and language modeling: report of a workshop held at the center for intelligent information retrieval, University of Massachusetts Amherst, September 2002, ACM SIGIR Forum 37 (1) (2003) 31-47.

[2]     A. Becks, S. Sklorz, M. Jarke, Exploring the Semantic Structure of Technical Document Collections, in: Proceedings Cooperative Information Systems, 7th International Conference (CoopIS 2000), Eilat, Israel, 2000, pp. 120-125.

[3]     J.M. Carmena, M.A. Lebedev, C.S. Henriquez, M.A.L. Nicolelis, Stable ensemble performance with single neuron variability during reaching movements in primates, Journal of Neuroscience 25 (46) (2005) 10712-10716.

[4]     S. Cheng, Y. Wang, Z. Wu, Provable Dimension Detection using Principal Component Analysis, in: Proceedings of the twenty-first annual symposium on Computational geometry, Pisa, Italy, 2005, pp. 208-217.

[5]     N. Choi, I. Song, H. Han, A Survey on Ontology Mapping, ACM SIGMOD Record, 35 (3) (2006), 34-41.

[6]     D. Cox, R. Savoy, Functional magnetic resonance imaging (fMRI) "brain reading": detecting and classifying distributed patterns of fMRI activity in human visual cortex, Neuroimage 19 (2003) 261-270.

[7]     S. Dasgupta, A. Gupta, An elementary proof of a theorem of Johnson and Lindenstrauss, Random Structures and Algorithms, 22 (1) (2003) 60 – 65.

4.     [8]     C. Ding, X. He, K-means Clustering via Principal Component Analysis, in: Proceedings of International Conference in Machine Learning (ICML 2004), Banff, Canada, 2004, pp. 225-232.

[9]     M. Dodge, Mapping the World-Wide Web, in: Preferred Placement: The Hit Economy, Hyperlink Diplomacy, and Web Epistemology, Symposium of the Design & Media Research Fellowship, Jan Van Eyck Akademie, Amsterdam, Netherlands, 1999, pp. 81-98.

[10]     J. Dolling, Commonsense Ontology and Semantics of Natural Language, Zeitschrift für Sprachtypologie und Universalienforschung (STUF) 46 (2) (1993) 133–141.

[11]     O. Etzioni, M. Cafarella, D. Downey, A. Popescu, T. Shaked, S. Soderland, D. Weld, A. Yates, Unsupervised named-entity extraction from the Web: An experimental study. Artificial Intelligence 165 (2005) 91-134.

[12]     C. Fellbaum, WordNet: An Electronic Lexical Database, The MIT Press, Cambridge, MA, USA, 1998.

[13]     J. Ford, F. Makedon, T. Steinberg, C. Owen, S. Johnson, A. Saykin, Stimulus tracking in Functional Magnetic Resonance Imaging (fMRI), in: Proceedings of the sixth ACM international conference on Multimedia, Bristol, UK, 1998, pp. 445-454.

[14]     J. Haynes, G. Rees, Predicting the stream of consciousness from activity in human visual cortex, Current Biology, 15 (2005) 1301-1307.

[15]     J. Haynes, G. Rees, Decoding mental states from brain activity in humans, Nature Reviews Neuroscience 7 (7) (2006) 523-534.

[16]     M. Hertzum, E. Frokjaer, Browsing and Querying in Online Documentation: A Study of User Interfaces and the Interaction Process, ACM Transactions on Computer-Human Interaction, 3 (2) (1996) 136-161.

[17]     Y. Kalfoglou, M. Schorlemmer, Ontology mapping: the state of the art, The Knowledge Engineering Review 18 (1) (2003) 1-31.

[18]     P. Kay, W. Kempton, What is the Sapir-Whorf Hypothesis? American Anthropologist 86 (1) (1984) 65-79.